\documentclass{article} 
\usepackage{PRIMEarxiv}

\usepackage{amsmath}
\usepackage{amsthm}
\usepackage{amsopn}
\usepackage{amsfonts}
\usepackage{amssymb}
\usepackage{mathrsfs}
\usepackage{bbm}
\usepackage{enumitem}
\usepackage{hyperref}
\usepackage{url}
\usepackage{graphicx}
\usepackage{subcaption}
\usepackage{algorithm}
\usepackage{algorithmic}
\usepackage{natbib}
\usepackage{booktabs}
\usepackage{cleveref}
\usepackage{todonotes}
\usepackage{pifont}

\usepackage{localmacros}

\newtheorem{remark}{Remark}

\title{Latent Diffeomorphic Dynamic Mode Decomposition}
\author{Willem Diepeveen \\
Department of Mathematics\\
University of California, Los Angeles\\
Los Angeles, CA 90095, USA \\
\texttt{wdiepeveen@math.ucla.edu} \\
\And
Jon Schwenk \\
Earth and Environmental Sciences Division \\
Los Alamos National Laboratory\\
Los Alamos, NM 87545, USA \\
\texttt{jschwenk@lanl.gov}
\And
Andrea L. Bertozzi \\
Department of Mathematics\\
University of California, Los Angeles \\
Los Angeles, CA 90095, USA \\
\texttt{bertozzi@ucla.edu}
}

\begin{document}

\maketitle

\begin{abstract}
    We present Latent Diffeomorphic Dynamic Mode Decomposition (LDDMD), a new data reduction approach for the analysis of non-linear systems that combines the interpretability of Dynamic Mode Decomposition (DMD) with the predictive power of Recurrent Neural Networks (RNNs). Notably, LDDMD maintains simplicity, which enhances interpretability, while effectively modeling and learning complex non-linear systems with memory, enabling accurate predictions. This is exemplified by its successful application in streamflow prediction.
\end{abstract}

\blfootnote{Our code is available at \href{https://github.com/wdiepeveen/Latent-Diffeomorphic-Dynamic-Mode-Decomposition}{https://github.com/wdiepeveen/Latent-Diffeomorphic-Dynamic-Mode-Decomposition}.}

\section{Introduction}
Predicting the next element in a sequence is rarely a straightforward, one-to-one process. Instead, it often relies on retaining and utilizing information from earlier in the sequence. For example, in language, the choice of the next word depends on more than just the preceding word. Similarly, in meteorology, identical weather conditions do not always lead to the same outcomes across different locations or times. Therefore, a crucial aspect of effective prediction is memory: the ability to learn and maintain relevant information over extended periods.

In many such applications, understanding the reasoning behind predictions -- that is, understanding what information has been learned and maintained -- is as important as making accurate forecasts. Streamflow prediction, crucial for flood preparedness and infrastructure design, illustrates this. Long Short-Term Memory networks (LSTMs) \cite{hochreiter1997long} -- a special type of Recurrent Neural Network (RNN) \cite{rumelhart1985learning} -- have revolutionized this field by outperforming traditional hydrological models for the streamflow prediction problem \cite{kratzert2018rainfall}. However, they often lack interpretability, which is a common issue among similar machine learning methods \cite{chung2015recurrent,haarnoja2016backprop}. 



In scenarios without memory where a one-to-one mapping exists, Dynamic Mode Decomposition (DMD) \cite{schmid2010dynamic} has proven highly effective as an interpretable data reduction method. Its success has led to numerous extensions \cite{tu2014on,dawson2016characterizing,hemati2017biasing}. 
Recent advancements have focused on adapting DMD to non-linear cases, primarily through Koopman operator theory \cite{Mezic2021}. This approach typically follows one of two paths: using non-linear dictionaries \cite{williams2015data,li2017extended}
or employing non-linear mappings \cite{takeishi2017learning,lusch2018deep,meng2024koopman,hou2024invertible}, with the latter being more popular. A key challenge in the non-linear mapping approach is the need for an (approximate) inverse function to recover the actual state from the dynamics in the mapped domain. While early attempts used approximations \cite{takeishi2017learning,lusch2018deep}, more recent research has shown that diffeomorphisms \cite{meng2024koopman,hou2024invertible} offer improved results.

Currently, all versions of Dynamic Mode Decomposition (DMD) rely on the assumption of a one-to-one mapping. This work aims to address this limitation by developing a novel approach that combines the high performance of RNN-like methods with the interpretability of DMD. Specifically, we aim to construct a non-linear DMD-like framework that mimics RNNs and can operate effectively in scenarios where the one-to-one mapping assumption does not hold. This would bridge the gap between the predictive power of RNN-like methods and the explanatory capabilities of traditional DMD, potentially opening up new avenues for interpretable machine learning for data reduction in complex, non-linear systems.


\paragraph{Contributions}
In this work we: (i) first propose Diffeomorphic Dynamic Mode Decomposition (DDMD) as a new non-linear DMD extension; (ii) then use it to construct Latent Diffeomorphic Dynamic Mode Decomposition (LDDMD) and discuss why this method effectively combines the strengths of DMD and RNNs while avoiding their respective limitations; and (iii) finally show that LDDMD shows promising results on toy and real data for streamflow prediction.

\vspace{-0.3cm}

\section{General Diffeomorphic Dynamic Mode Decomposition}
Dynamic Mode Decomposition (DMD) assumes that the evolution of a time series $\{\Vector^\sumIndA\}_{\sumIndA=0}^\numData \subset \Real^\dimInd$ is governed by an unknown matrix $\dynGen \in \Real^{\dimInd \times \dimInd}$ through
\begin{equation}
    \Vector^{\sumIndA} = \dynGen \Vector^{\sumIndA-1}, \quad \sumIndA = 1, \ldots, \numData.
    \label{eq:standard-dmd-evolution}
\end{equation} 
In addition, assuming that $\dimInd$ is even, and $\dynGen - \mathbf{I}$ is diagonalizable with all eigenvalues having a non-zero imaginary component, we can always write (\ref{eq:standard-dmd-evolution}) as
\begin{equation}
    \Vector^{\sumIndA} = \linInv^{-1} \dynGenLatent \linInv \Vector^{\sumIndA-1}, \quad \sumIndA = 1, \ldots, \numData,
    \label{eq:diagonal-dmd-evolution}
\end{equation}
where $\linInv \in \Real^{\dimInd \times \dimInd}$ is an invertible matrix and $\dynGenLatent \in \Real^{\dimInd \times \dimInd}$ is a block diagonal matrix with $\dimInd/2$ blocks of the form
\begin{equation}
    e^{-\mu_\sumIndB \Delta t}\left[\begin{array}{ll}
    \cos (\networkParamsC_{\sumIndB} \Delta t) & -\sin (\networkParamsC_{\sumIndB} \Delta t) \\
    \sin (\networkParamsC_{\sumIndB} \Delta t) & \cos (\networkParamsC_{\sumIndB} \Delta t)
    \end{array}\right], \quad \mathrm{for} \; \sumIndB = 1, \ldots, \dimInd/2,
    \label{eq:K-blocks}
\end{equation}
with $\mu_\sumIndB, \networkParamsC_{\sumIndB} \in \Real$ and $\Delta t >0$ being the time between observations in the time series. 

For the above assumptions on $\dynGen$ to hold, the time series needs (at least) to be centered, i.e., $\frac{1}{\numData+1} \sum_{\sumIndA=0}^\numData \Vector^\sumIndA = \mathbf{0}$. For a non-centered time series, we can repeat the above for $\{\Vector^\sumIndA - \linBias\}_{\sumIndA=0}^\numData$ with $\linBias = \frac{1}{\numData+1} \sum_{\sumIndA=0}^\numData \Vector^\sumIndA$, which gives
\begin{equation}
    \Vector^{\sumIndA} = \linInv^{-1} \dynGenLatent \linInv (\Vector^{\sumIndA-1} - \linBias) + \linBias, \quad \sumIndA = 1, \ldots, \numData.
    \label{eq:diagonal-dmd-evolution}
\end{equation}

We propose \emph{Diffeomorphic Dynamic Mode Decomposition (DDMD)}, which directly extends (\ref{eq:diagonal-dmd-evolution}) by replacing the invertible linear mapping $\Vector \mapsto \linInv(\Vector - \linBias)$ by a diffeomorphism $\diffeo:\Real^\dimInd \to \Real^\dimInd$. In particular, DDMD assumes that
\begin{equation}
    \Vector^{\sumIndA} = (\diffeo^{-1} \circ \dynGenLatent \circ \diffeo) (\Vector^{\sumIndA-1}) = (\diffeo^{-1} \circ (\dynGenLatent)^\sumIndA \circ \diffeo) (\Vector^{0}), \quad \sumIndA = 1, \ldots, \numData.
    \label{eq:diffeomorphic-diagonal-dmd-evolution}
\end{equation}

\begin{remark}
    We note that this approach can be seen as: (i) a combination of the approaches taken in \cite{lusch2018deep} -- where a similar parametrization for $\dynGenLatent$ is used, but $\diffeo^{-1}$ and $\diffeo$ are replaced by (non-invertible) neural networks -- and state-of-the-art approaches taken in \cite{meng2024koopman,hou2024invertible} -- where diffeomorphisms are used, but no block diagonal assumptions are made on $\dynGenLatent$, \revA{and (ii) a special case of \cite{Bollt2018} where the eigenvectors of the Koopman operator are only assumed to generate a local diffeomorphism}.
\end{remark}


\section{Latent Diffeomorphic Dynamic Mode Decomposition}
The actual problem we are interested is an adaptation of the DDMD problem. That is, given a time series $\{\Vector^\sumIndA\}_{\sumIndA=0}^\numData \subset \Real^\dimInd$, we aim to predict the corresponding sequence element in $\{ \corVector^\sumIndA\}_{\sumIndA=0}^\numData \subset \Real^{\dimIndB}$, assuming that there is no one-to-one mapping $\Vector \mapsto \corVector$, i.e., there is memory in the system. To address the one-to-one mapping assumption still inherent in DDMD, we introduce additional variables to predict the next sequence element. In particular, we assume that the evolution of a time series $\{(\Vector^\sumIndA, \corVector^\sumIndA)\}_{\sumIndA=0}^\numData \subset \Real^\dimInd \times \Real^{\dimIndB}$ is governed by dynamics 
\begin{equation}
    (\Vector^{\sumIndA}, \latentVector^{\sumIndA}) = (\Phi^{-1} \circ \mathcal{K} \circ \Phi) (\Vector^{\sumIndA-1}, \latentVector^{\sumIndA-1}), \quad \corVector^{\sumIndA} = g(\latentVector^{\sumIndA}).
    \label{eq:full-latent-diffeomorphic-diagonal-dmd-evolution}
\end{equation}
Here, $\Phi: \Real^\dimInd \times \Real^{\dimIndC}\to \Real^\dimInd \times \Real^{\dimIndC}$ is a diffeomorphism defined as
\begin{equation}
    \Phi(\Vector, \latentVector) := (\diffeo_1(\Vector), \coupling(\Vector) + \diffeo_2(\latentVector)),
\end{equation}
where $\diffeo_1: \Real^\dimInd \to \Real^\dimInd$ and $\diffeo_2: \Real^\dimIndC \to \Real^\dimIndC$ are diffeomorphisms and $\coupling: \Real^\dimInd \to \Real^\dimIndC$ is a coupling mapping. The inverse of $\Phi$ is given by
\begin{equation}
    \Phi^{-1}(\VectorB, \latentVectorB) = (\diffeo_{1}^{-1}(\VectorB), \diffeo_{2}^{-1}(\latentVectorB - \coupling(\diffeo_{1}^{-1}(\VectorB)))).
\end{equation}
Furthermore, 
\begin{equation}
    \mathcal{K} = \begin{bmatrix}
\dynGenLatent_1 &  \\
 & \dynGenLatent_2 \\
\end{bmatrix},
\end{equation}
where
$\dynGenLatent_1: \Real^{\dimInd \times\dimInd}$ and $\dynGenLatent_2: \Real^{\dimIndC \times\dimIndC}$ are block-diagonal matrices with blocks of the form (\ref{eq:K-blocks})\footnote{Note that for this to be well-defined, we need both $\dimInd$ and $\dimIndC$ to be even.}, $g: \Real^\dimIndC\to \Real^\dimIndB$ is a mapping, and the latent variable $\latentVector$ is initialized by some $\latentVector^0 \in \Real^\dimIndC$.

\begin{remark}
    The dynamical system (\ref{eq:full-latent-diffeomorphic-diagonal-dmd-evolution}) encodes a state defined by $\Vector$ and $\latentVector$, where $\Vector$ is independent of $\latentVector$, but $\latentVector$ is partially driven by $\Vector$, and $\corVector$ depends on $\latentVector$, which incorporates both $\Vector$ and system memory, thus breaking the one-to-one assumption of previous DMD versions.
\end{remark}

The goal of \emph{Latent Diffeomorphic Dynamic Mode Decomposition (LDDMD)} is slightly different than  DDMD. That is, we are not interested in explicitly representing the underlying dynamics of the $\Vector^\sumIndA$, but want to predict $\corVector^\sumIndA$ given $\Vector^\sumIndA$ and the current time index $\sumIndA$. In other words, we aim to represent the dynamics of the $\latentVector^{\sumIndA}$, which allows us to focus on the $\latentVector^{\sumIndA}$ part in (\ref{eq:full-latent-diffeomorphic-diagonal-dmd-evolution}), i.e.,
\begin{equation}
    \latentVector^{\sumIndA} = \diffeo_{2}^{-1}(\dynGenLatent_2 (\coupling(\Vector^{\sumIndA-1}) + \diffeo_2(\latentVector^{\sumIndA-1})) - \coupling(\Vector^{\sumIndA})) = \diffeo_{2}^{-1}((\dynGenLatent_2)^{\sumIndA} (\coupling(\Vector^0) + \diffeo_2(\latentVector^0)) - \coupling(\Vector^{\sumIndA})).
    \label{eq:almost-latent-diffeomorphic-diagonal-dmd-evolution}
\end{equation}
Since $\latentVector^0$ is some initialization, we are free to write $\latentVectorB_0$ for $\coupling(\Vector^0) + \diffeo_2(\latentVector^0)$. Also dropping all the subscripts gives us the LDDMD problem, where we assume that
\begin{equation}
    \corVector^{\sumIndA} = g(\latentVector^{\sumIndA}) =  g(\diffeo^{-1}((\dynGenLatent)^{\sumIndA} \latentVectorB_0 - \coupling(\Vector^{\sumIndA}))),
    \label{eq:latent-diffeomorphic-diagonal-dmd-evolution}
\end{equation}
and aim to learn the mapping $g$, the diffeomorphism $\diffeo$, the coupling $\coupling$, and the initialization $\latentVectorB^0$ along with the $\mu_\sumIndB, \networkParamsC_{\sumIndB}$ that parametrize $\dynGenLatent$ through (\ref{eq:K-blocks}).

\begin{remark}
    The latent dynamics $(\dynGenLatent)^{\sumIndA} \latentVectorB^0$ encode the memory of the system and is evolving independently of $\Vector^\sumIndA$. In other words, we really have a non-one-to-one mapping from $\Vector^\sumIndA$ to $\corVector^{\sumIndA}$, where the memory is encoded in a more interpretable way than for RNN-like methods, with the key difference that we explicitly have to solve for an initial condition $\latentVectorB_0$. Having said that, standard RNNs famously have trouble with maintaining information for longer terms and can be unstable. Both of these issues can be attributed to vanishing or exploding gradients, which LDDMD can also suffer from for $\mu_\sumIndB \neq 0$ \revA{in $\dynGenLatent$, see (\ref{eq:K-blocks})}. So in practice, we assume for LDDMD that $\mu_\sumIndB = 0$.
\end{remark}


\section{Training}
To learn an LDDMD representation of a data set $\{(\Vector^\sumIndA, \corVector^\sumIndA)\}_{\sumIndA=0}^\numData \subset \Real^\dimInd \times \Real^{\dimIndB}$, we need loss functions, parametrizations and initializations thereof.

We propose the following training loss to solve for the LDDMD approximation of a time series $\{(\Vector^\sumIndA, \corVector^\sumIndA)\}_{\sumIndA=0}^\numData$ with an even dimension $\dimIndC$ for the latent space:
\begin{equation}
    \mathcal{L}_{LDDMD} (\networkParams, \networkParamsB,  \networkParamsC, \latentVectorB_0, \networkParamsD) := \sum_{\sumIndA = 0}^\numData \|\corVector^{\sumIndA} - g_{\networkParamsD}(\diffeo_{\networkParams}^{-1}((\dynGenLatent_{\networkParamsC})^{\sumIndA} \latentVectorB_{0} - \coupling_\networkParamsB(\Vector^{\sumIndA})))\|_2.
\end{equation}

Next, although the specific neural networks used to parametrize the mappings might vary from application to application, there are some general choices for the case $\dimIndB=1$ -- which is the case for streamflow prediction -- that come with well-motivated initializations.

In this case, we found the following parameterization suitable: (i) the diffeomorphism as an invertible neural network composed of a single additive coupling layer \cite{dinh2014nice} with learnable polynomial activation on either the even or odd indices, initialized such that $\diffeo_{\networkParams^{(0)}}(\Vector) = \Vector$; (ii) the coupling as a feedforward network with learnable polynomial activation functions, initialized such that $\coupling_{\networkParamsB^{(0)}}(\Vector) \approx \mathbf{0}$; (iii) the frequencies $\networkParamsC_\sumIndB$ as learnable parameters, initialized as the $\dimIndC/2$ most prevalent frequencies in the Fourier spectrum\footnote{Factoring in the time $\Delta t$ between observations.} of the 1D signal $\{\corVector^\sumIndA\}_{\sumIndA=0}^\numData$; (iv) the latent initialization as a learnable vector, initialized as $\latentVectorB^{(0)}_0 = \mathbf{0}$; (v) and the mapping $g_{\networkParamsD}$ as a feedforward network with softplus activation functions and random initialization. \revA{Limiting the diffeomorphism to only use additive coupling layers serves as a form of regularization, helping to prevent the mapping from producing excessively large or small output values.}


\section{Numerical experiments}
\label{sec:numerics}
In this section, we demonstrate the performance of LDDMD through experiments on data reduction for streamflow data $\{\corScalar^\sumIndA\}_{\sumIndA=0}^\numData \subset \Real$ from synthetic and real meteorological data $\{\Vector^\sumIndA\}_{\sumIndA=0}^\numData \subset \Real^\dimInd$. Specifically, we show that LDDMD effectively reduces the dataset using a low-dimensional latent space and extrapolates beyond its training data, both in nearly ideal settings with noisy data and in real-world scenarios.

The key metric that will be used to quantify performance is the \emph{Nash–Sutcliffe model efficiency coefficient (NSE)}:
\begin{equation}
    \mathrm{NSE}=1-\frac{\sum_{\sumIndA = 0}^\numData (\corScalar^{\sumIndA} - g_{\networkParamsD}(\diffeo_{\networkParams}^{-1}((\dynGenLatent_{\networkParamsC})^{\sumIndA} \latentVectorB_{0} - \coupling_\networkParamsB(\Vector^{\sumIndA}))) )^2}{\sum_{\sumIndA = 0}^\numData (\corScalar^{\sumIndA} - \bar{\corScalar})^2},
\end{equation}
where $\bar{\corScalar} := \frac{1}{\numData}\sum_{\sumIndA = 0}^\numData \corScalar^{\sumIndA}$. For intuition, a higher NSE is better, where $\mathrm{NSE} = 1$ is perfect and a model having $\mathrm{NSE} = 0.5$ or over indicates that the dynamics are captured well. For reference, we also compare our results to LSTMs using the standard implementation in \texttt{NeuralHydrology} \cite{kratzert2022neuralhydrology}, which is a widely used package that is optimized for streamflow prediction.

For all experiments in this section, details on the data sets are provided in \emph{Supplementary Material 1}, detailed LDDMD-training configurations in \emph{Supplementary Material 2} and additional results in \emph{Supplementary Material 3}.

Finally, all of the experiments are implemented using \texttt{PyTorch} in Python 3.8 and run on a 2 GHz Quad-Core Intel Core i5 with 16 GB RAM.



\begin{figure}[h!]
    \centering
    \begin{subfigure}[b]{0.49\textwidth}
        \centering
        \includegraphics[width=0.77\textwidth]{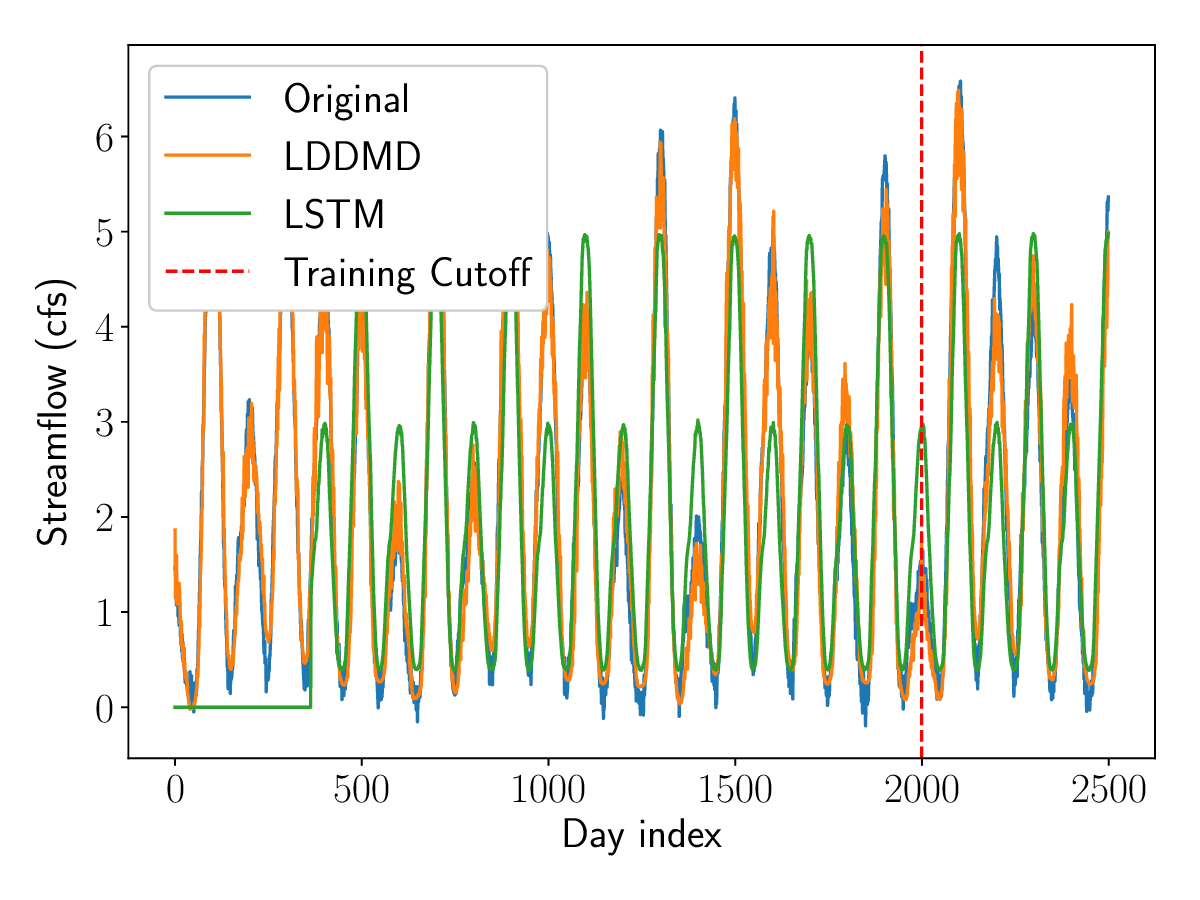}
        \caption{Synthetic data results}
        \label{fig:toy-reconstruction}
    \end{subfigure}
    \hfill
    \begin{subfigure}[b]{0.49\textwidth}
        \centering
        \includegraphics[width=0.77\textwidth]{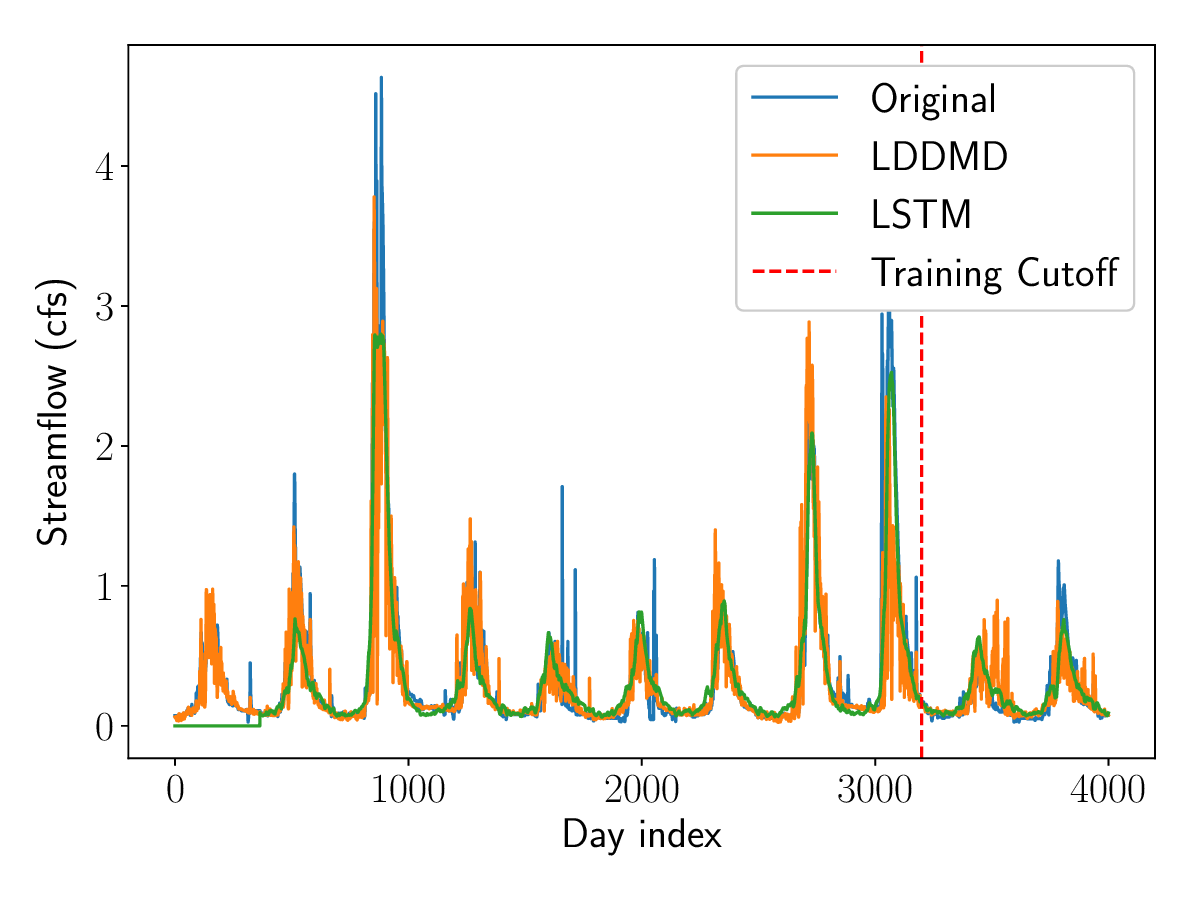}
        \caption{Real-world data results}
        \label{fig:crb-reconstruction}
    \end{subfigure}
    \caption{For both synthetic and real-world data LDDMD (proposed) is not only able to learn the underlying dynamics from the training data, but also to extrapolate far beyond the training horizon. The \texttt{NeuralHydrology} LSTM model performs well on data that it is optimized for (b), but is doing worse on non-typical streamflow data (a).}
    \label{fig:reconstruction-results}
\end{figure}

\paragraph{Synthetic data}
For the synthetic data set, 2D data and a 2D latent process govern the streamflow through dynamics of the form (\ref{eq:full-latent-diffeomorphic-diagonal-dmd-evolution}). The data used for training the LDDMD model comes from this underlying dynamics model, but noise has been added to it. The dimensions $\dimInd=2$ and $\dimIndC=2$ are also used for parametrizing the mappings in the LDDMD model.
The results in Figure~\ref{fig:toy-reconstruction} suggest that despite the noise we are able to find a model to fit the data, predict the signal well-beyond its training window and outperform the LSTM: for LDDMD we have $\mathrm{NSE} = 0.96$ on training data and $\mathrm{NSE} = 0.94$ on validation data, whereas for the LSTM\footnote{We do not take the 365 day sequence length of the LSTM into account here for the NSE.} we have $\mathrm{NSE} = 0.80$ on training data and $\mathrm{NSE} = 0.80$ on validation data. 
Notably, the model has arguably almost recovered the ground truth. In particular, the learned frequency $\networkParamsC_1 = 0.0097$ is almost exactly the ground truth frequency $\networkParamsC^* = 0.0099$ and despite over-parametrization of the coupling mapping $\coupling_\networkParamsB$, even the latent dynamics look similar to the ground truth up to rigid body transformation and rescaling\footnote{Note that this is not expected at all as there is just a 1-dimensional variable $\corScalar^\sumIndA$ to reconstruct each two-dimensional variable $\latentVector^\sumIndA$, but the fact that we are able to get a similar shape reminisces of Takens' theorem \cite{takens2006detecting}. We leave exploration of this connection for future research.}, from which we retrieve the information that the the process that governs the streamflow oscillates between two regions of the latent space (see Figure~2 in \emph{Supplementary Material 3}). 

\paragraph{Real-world data}
For the real-world data, we have $\dimInd=14$ for the meteorological data and use $\dimIndC=10$ for the latent space to approximate the streamflow. We parametrize the mappings in the LDDMD model accordingly.
The results in Figure \ref{fig:crb-reconstruction} indicate that the model effectively fits the data, albeit with lower performance compared to the numericaly optimized LSTM, and provides (almost) practically useful streamflow predictions that extend well beyond the training window: for LDDMD we have $\mathrm{NSE} = 0.75$ on training data and $\mathrm{NSE} = 0.46$ on validation data, whereas for the LSTM we have $\mathrm{NSE} = 0.89$ on training data and $\mathrm{NSE} = 0.89$ on validation data. 
Despite the performance gap between our general-purpose LDDMD and the numerically optimized LSTM in terms of NSE, the results remain encouraging and highlight key areas for further improvement of the base method. Although a thorough analysis of the performance gap is left for future research, we expect that possible explanations involve that 
LSTMs are better at removing noise from the signal (see Figure~3 in \emph{Supplementary Material 3}), which would significantly increase the NSE. Finally, as there is no ground truth process available, we also leave matching these results to possible underlying processes for future research. 




\section{Conclusions}
\vspace{-0.2cm}
In conclusion, this work presents Latent Diffeomorphic Dynamic Mode Decomposition (LDDMD), a novel method that combines Dynamic Mode Decomposition and Recurrent Neural Networks for data reduction and nonlinear time series prediction. LDDMD offers both interpretability and strong performance, as shown in streamflow forecasting, though further improvements are needed for state-of-the-art results.


\vspace{-0.4cm}
\section*{Acknowledgments}
\vspace{-0.2cm}
All authors are supported by the U.S. Department of
Energy, Office of Science, Office of Advanced Scientific Computing Research
under award DE-SC0025589 and under Triad National Security, LLC (‘Triad’) contract grant
89233218CNA000001 [FWP: LANLE2A2].
JS is also supported by the Laboratory Directed Research and Development program of Los Alamos National Laboratory under project numbers 20170668PRD1 and 20210213ER.
ALB is also supported by NSF grant DMS-2152717.

\vspace{-0.4cm}
\bibliographystyle{plain}
\bibliography{bibliography}

\appendix
\section{Data set details}

\subsection{Synthetic data}
The synthetic data in Section 5 of the main article is of the form 
$$
    (\Vector^{\sumIndA}, \latentVector^{\sumIndA}) = (\Phi^{-1} \circ \mathcal{K} \circ \Phi) (\Vector^{\sumIndA-1}, \latentVector^{\sumIndA-1}), \quad \corVector^{\sumIndA} = g(\latentVector^{\sumIndA}).
$$
with
$$
    \Phi(\Vector, \latentVector) := (\diffeo_1(\Vector), \coupling(\Vector) + \diffeo_2(\latentVector)),
$$
and
$$
    \mathcal{K} = \begin{bmatrix}
\dynGenLatent_1 &  \\
 & \dynGenLatent_2 \\
\end{bmatrix}.
$$
The diffeomorphisms are given by
$$
    \diffeo_1(\Vector) := (2(\Vector_1 - \sin(\Vector_2)), \frac{1}{4} \Vector_2 ),
$$
and
$$
    \diffeo_2(\Vector) := (2(\Vector_1 - \Vector_2^2 - 3), 3\Vector_2 ),
$$
the coupling is given by
$$
    \coupling(\Vector) := (\Vector_1^2 + \Vector_2^2, \Vector_1 - \Vector_2),
$$
and the matrices $\dynGenLatent_1$ and $\dynGenLatent_2$ are parametrized as
$$
    e^{-\mu_\sumIndB \Delta t}\left[\begin{array}{ll}
    \cos (\networkParamsC_{\sumIndB} \Delta t) & -\sin (\networkParamsC_{\sumIndB} \Delta t) \\
    \sin (\networkParamsC_{\sumIndB} \Delta t) & \cos (\networkParamsC_{\sumIndB} \Delta t)
    \end{array}\right],
$$
with $\mu_1 = \mu_2 = 0$, $\networkParamsC_{1} = \frac{\pi}{100}$, $\networkParamsC_{2} = \frac{\pi}{100 \sqrt{10}}$ and $\Delta t = 1$. 

The system is initialized as 
$$
    (\Vector^0, \latentVector^0) := \Phi^{-1}([0,1]^\top, [1,1]^\top).
$$

Finally, 
$$
    g(\latentVector) := \operatorname{softplus}(- \latentVector_1 - \frac{3}{4}\latentVector_2 + 1\frac{1}{2}).
$$

\begin{figure}[h!]
    \centering
    \begin{subfigure}[b]{0.49\textwidth}
        \centering
        \includegraphics[width=\textwidth]{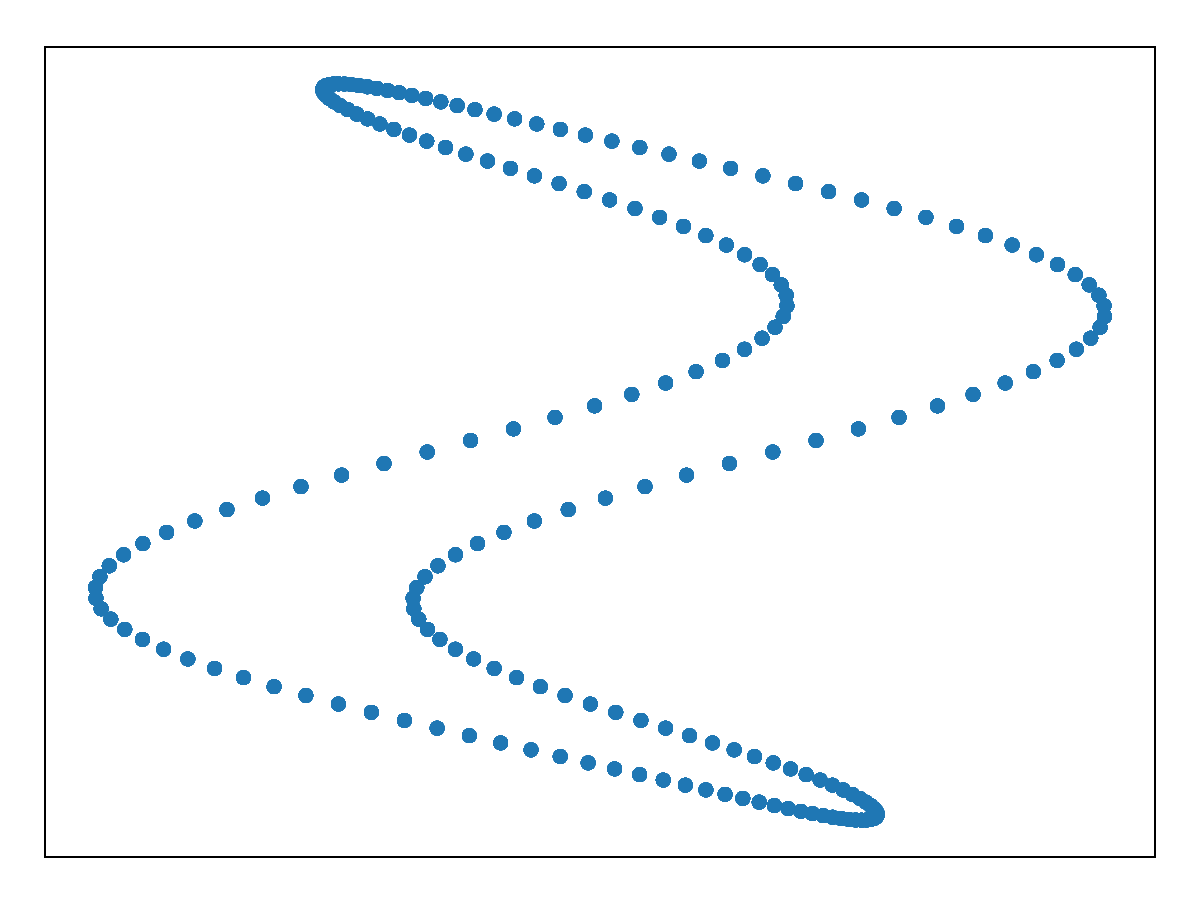}
        \caption{Clean $\Vector$-variable data}
    \end{subfigure}
    \hfill
    \begin{subfigure}[b]{0.49\textwidth}
        \centering
        \includegraphics[width=\textwidth]{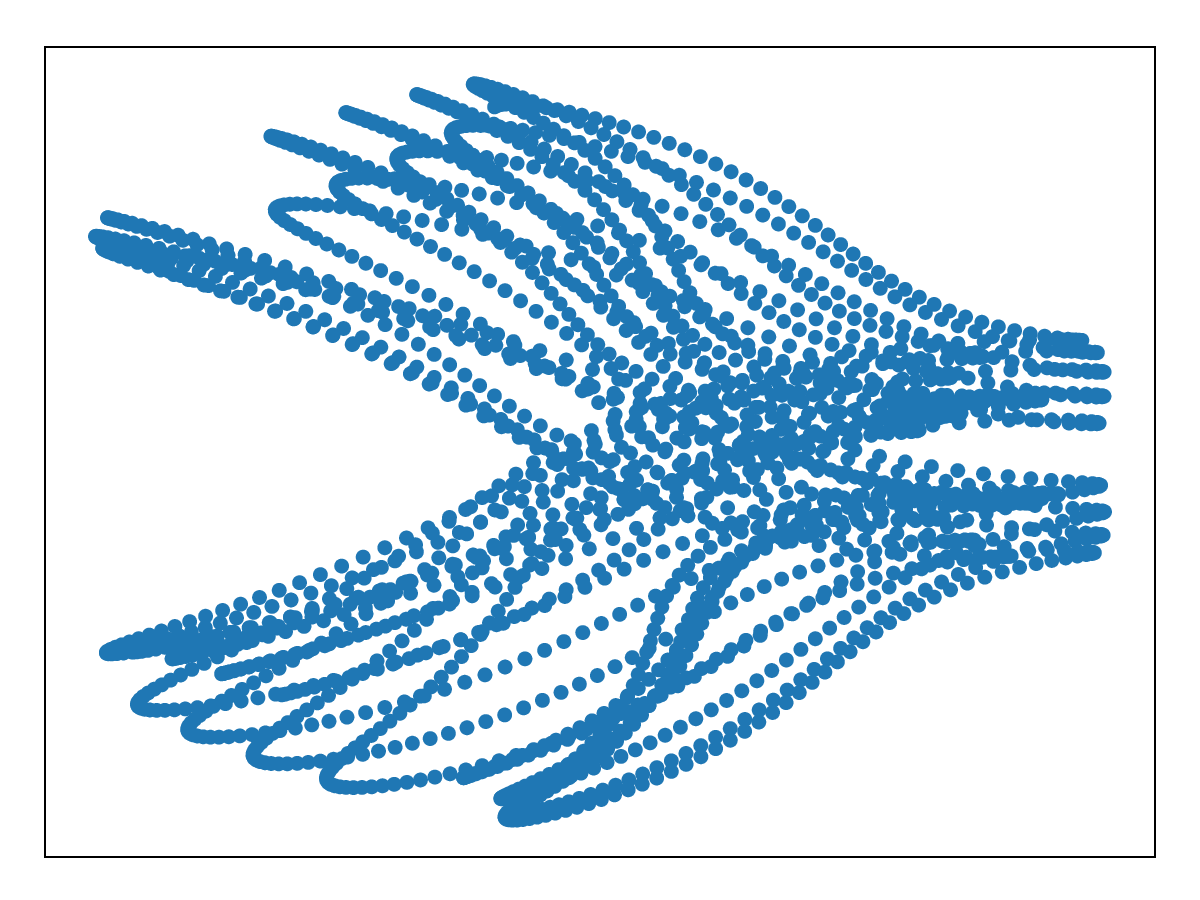}
        \caption{Clean $\latentVector$-variable data}
    \end{subfigure}
    \caption{A visualization of the synthetic data: whereas the $\Vector$-variable follows a periodic orbit, the latent $\latentVector$-variable data are much more complicated due to the interplay with the memory of the system.}
    \label{fig:data}
\end{figure}

As the $\Vector$ and $ \latentVector$-variables are 2-dimensional, we can visualize them separately (see Figure \ref{fig:data}).

\subsection{Real data}


Streamflow observations ($\corVector$ variable) were provided by the United States Geological Survey (USGS). The example here is from Muddy Creek near Emery, Utah (USGS id 09330500). These data are available at \href{https://waterdata.usgs.gov/monitoring-location/09330500/#dataTypeId=continuous-00065-0&period=P7D}{https://waterdata.usgs.gov/monitoring-location/09330500/\#dataTypeId=continuous-00065-0\&period=P7D}. Streamflow data are provided as daily averages in cubic feet per second. These data were normalized by the drainage area of the watershed of the gage, or 271.9 km$^2$ in this case, and converted to millimeters per day to arrive at $\corVector$.

The $\Vector$-variables were sampled over the watershed of the Muddy Creek gage by spatial averaging (or summing in the case of total precipitation). These variables include:

\begin{table}[h!]
    \centering
    \label{tab:meteo-vars}
    \footnotesize
    \begin{tabular}{l|l}
         \hline
\textbf{Variable Name} & \textbf{Variable Meaning}  \\ \hline
dewpoint\_temperature\_2m\_\_mean\_\_era5l\_daily & Daily mean dewpoint temperature at 2 meters  \\ \hline
potential\_evaporation\_\_sum\_\_era5l\_daily & Daily sum of potential evaporation \\ \hline
snow\_depth\_water\_equivalent\_\_mean\_\_era5l\_daily & Daily mean snow depth water equivalent  \\ \hline
surface\_net\_solar\_radiation\_\_mean\_\_era5l\_daily & Daily mean surface net solar radiation \\ \hline
surface\_net\_thermal\_radiation\_\_mean\_\_era5l\_daily & Daily mean surface net thermal radiation  \\ \hline
surface\_pressure\_\_mean\_\_era5l\_daily & Daily mean surface atmospheric pressure \\ \hline
temperature\_2m\_\_mean\_\_era5l\_daily & Daily mean air temperature at 2 meters \\ \hline
total\_precipitation\_\_sum\_\_era5l\_daily & Daily sum of total precipitation \\ \hline
u\_component\_of\_wind\_10m\_\_mean\_\_era5l\_daily & Daily mean east-west wind component at 10 meters \\ \hline
v\_component\_of\_wind\_10m\_\_mean\_\_era5l\_daily & Daily mean north-south wind component at 10 meters \\ \hline
volumetric\_soil\_water\_layer\_1\_\_mean\_\_era5l\_daily & Daily mean volumetric soil water content, Layer 1 (0–7 cm) \\ \hline
volumetric\_soil\_water\_layer\_2\_\_mean\_\_era5l\_daily & Daily mean volumetric soil water content, Layer 2 (7–28 cm) \\ \hline
volumetric\_soil\_water\_layer\_3\_\_mean\_\_era5l\_daily & Daily mean volumetric soil water content, Layer 3 (28–100 cm) \\ \hline
volumetric\_soil\_water\_layer\_4\_\_mean\_\_era5l\_daily & Daily mean volumetric soil water content, Layer 4 (100–289 cm) \\ \hline
    \end{tabular}
\end{table}


\clearpage
\section{Training details}
\revA{Based on our experience, both shallow and deep networks delivered strong performance, provided the activation functions were regular. For instance, non-regular ReLU activation performed poorly. We prioritized simplicity in our models, which primarily guided the selection of the parameters reported below.}
\paragraph{Common parameters}
\begin{itemize}
    \item \textbf{Batch size}: 256
    \item \textbf{Optimizer}: Adam with \texttt{betas} = (0.9, 0.99) and learning rate $10^{-3}$.
    \item \textbf{Model Architecture}:
    \begin{itemize}
        \item \textbf{Diffeomorphisms}: 
        Additive coupling layer \cite{dinh2014nice}, which adds the mapping of the sum of two adjacent parity-0 inputs through a learnable order-2 polynomials to corresponding parity-1 entries (i.e., the parity-1 entry in between the two parity-0 entries).
        \item \textbf{Couplings}: Multi layer perceptron network with $\ell_f$ hidden layers with dimension $\dimInd_f$ (different per data set) and learnable polynomial activation functions of order 2 (unique polynomial per neuron).
        \item \textbf{Mappings}: Multi layer perceptron with one hidden layer with dimension $\dimInd_g$ and softplus activation.
    \end{itemize}
\end{itemize}

\paragraph{Data set-specific parameters}
The remaining parameters are summarized below:
\begin{table}[h!]
    \centering
    \begin{tabular}{l|l|l|l|l}
         \hline
\textbf{Data set} & $\ell_f$ & $\dimInd_f$ & $\dimInd_g$ & \textbf{Epochs}  \\ \hline
Synthetic & 2 & 2 & 4 & 1000 \\ \hline
Real & 1 & 40 & 4 & 200 \\ \hline \hline
    \end{tabular}
\end{table}

\clearpage

\section{Additional numerical results} 

\begin{figure}[h!]
    \centering
    \includegraphics[width=0.49\linewidth]{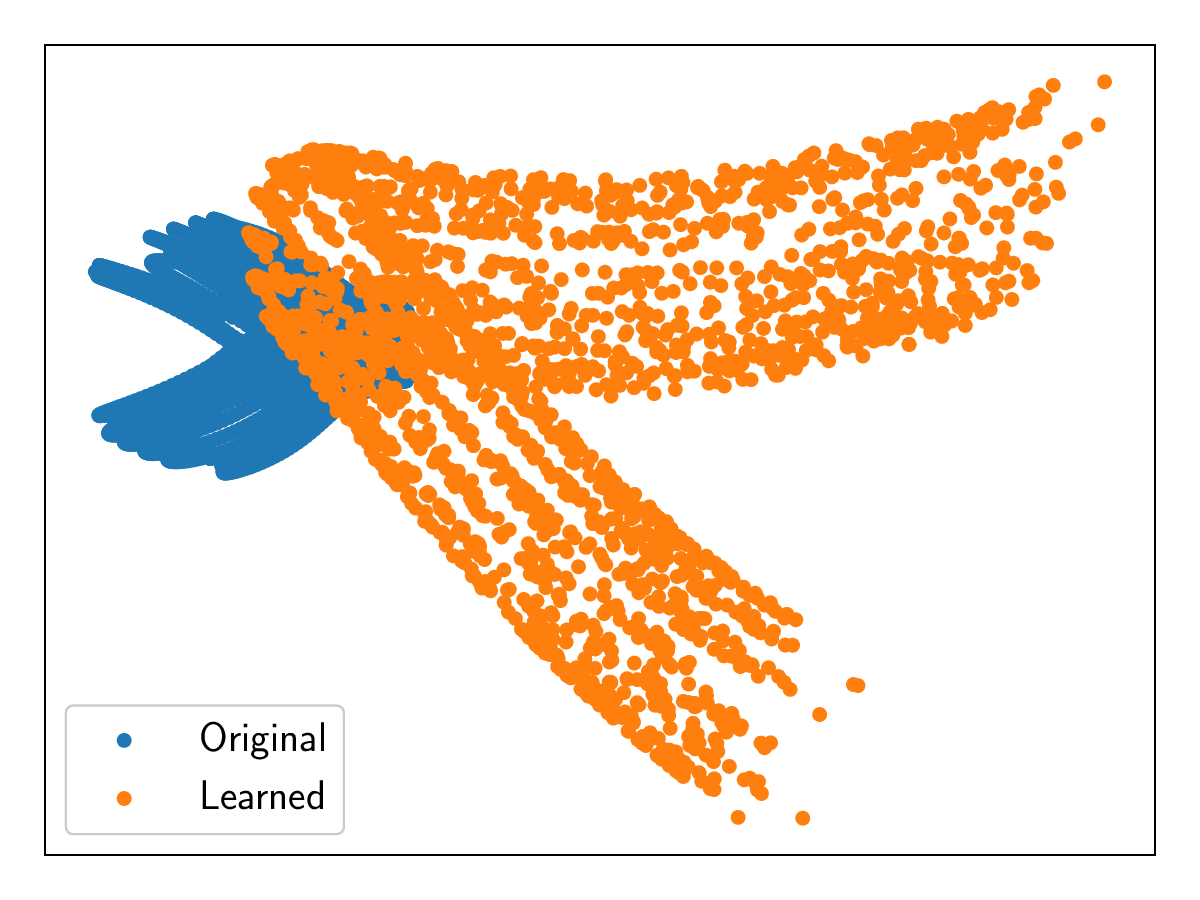}
    \caption{The ground truth and learned latent ($\latentVector$-variable) dynamics look very similar, but differ roughly by a rigid body transformation and rescaling. This indicates that the latent space does carry information for the synthetic data, i.e., LDDMD has an interpretable latent space.}
\end{figure}


\begin{figure}[h!]
    \centering
    \begin{subfigure}[b]{0.49\textwidth}
        \centering
        \includegraphics[width=0.95\textwidth]{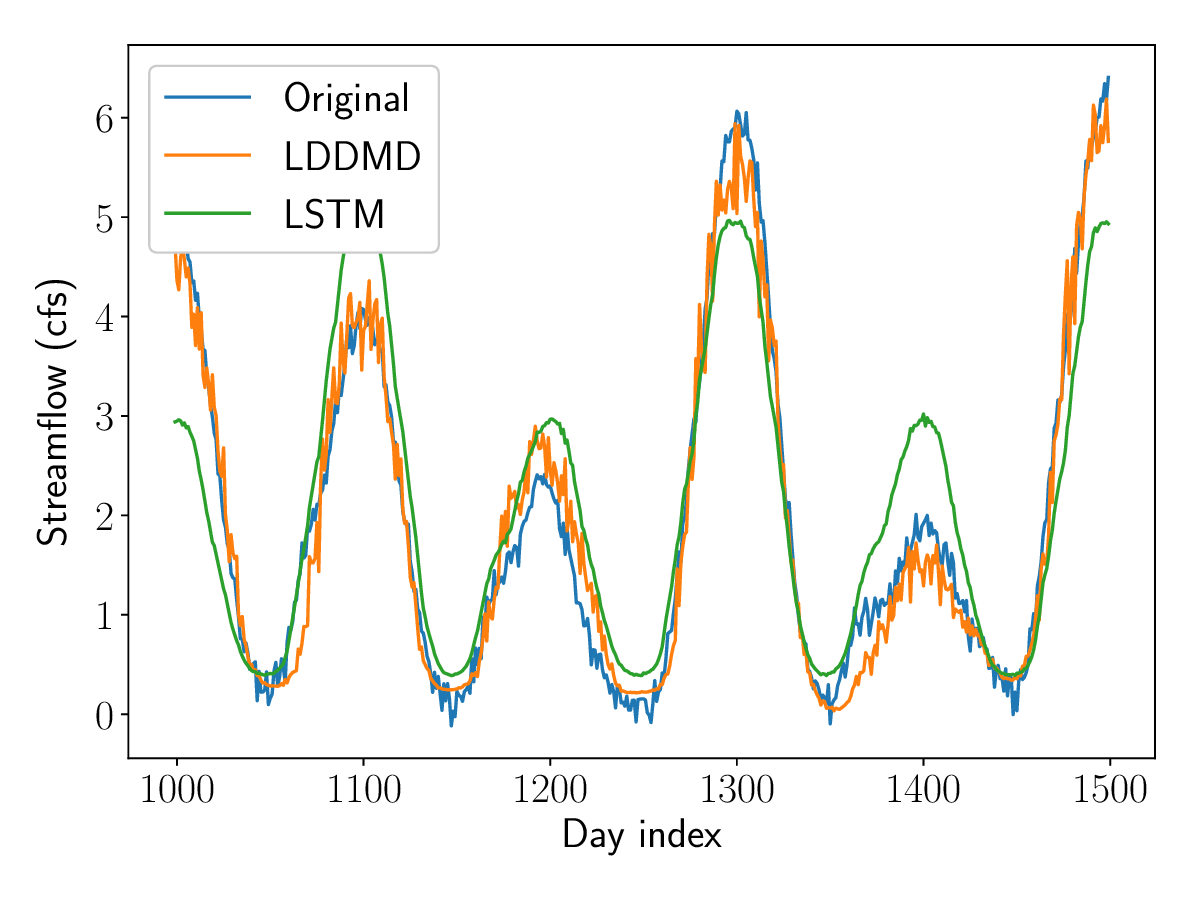}
        \caption{Syntethic data results}
        \label{fig:toy-reconstruction}
    \end{subfigure}
    \hfill
    \begin{subfigure}[b]{0.49\textwidth}
        \centering
        \includegraphics[width=0.95\textwidth]{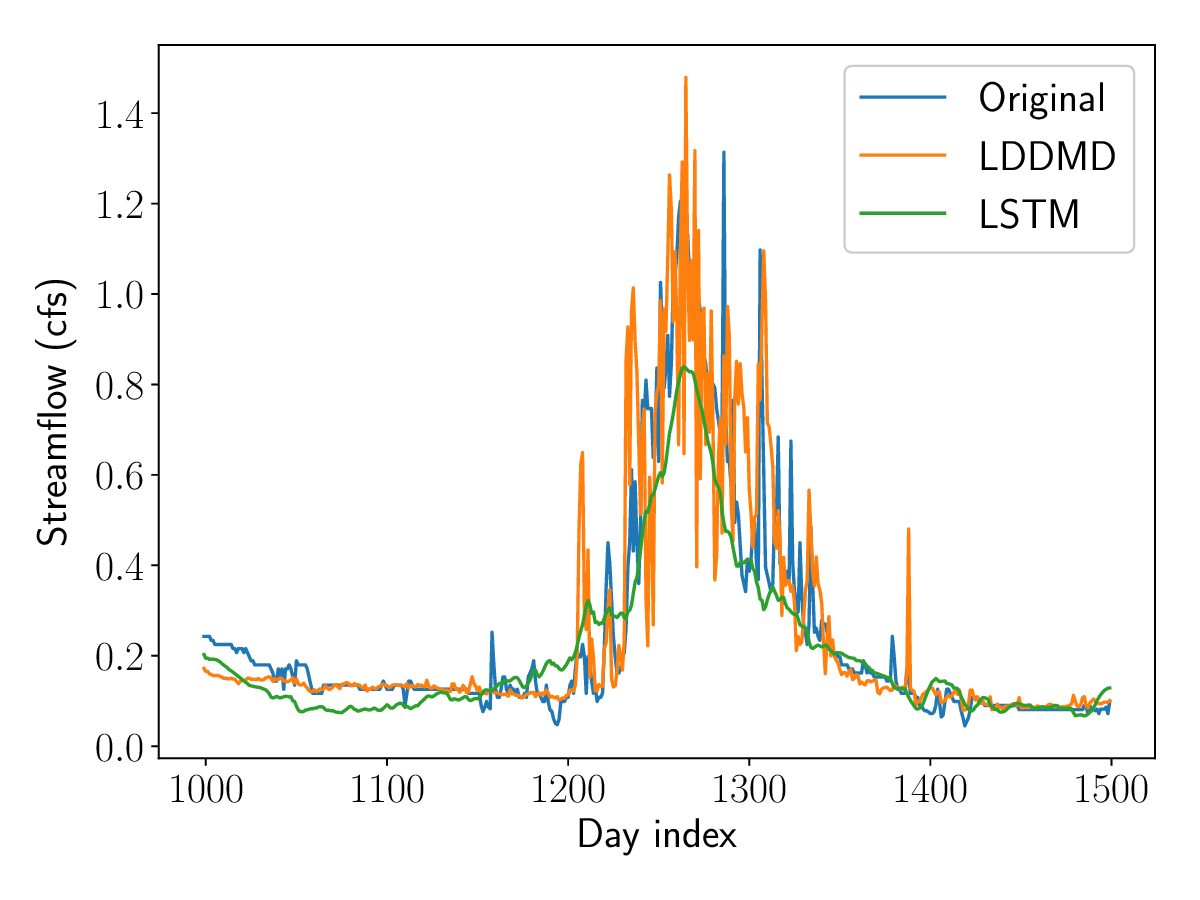}
        \caption{Real-world data results}
        \label{fig:crb-reconstruction}
    \end{subfigure}
    \caption{When examining a specific interval in the plots in Figure~1, it is evident that the LSTM predictions are considerably less noisy than those of LDDMD. This difference is likely due to the inherent averaging in LSTMs, where predictions at each time step incorporate information from many preceding time points, effectively smoothing out noise. In contrast, LDDMD lacks a comparable denoising mechanism.}
    \label{fig:reconstruction-results}
\end{figure}

\end{document}